\documentclass[a4paper, 10pt, conference]{ieeeconf}      % Use this line for a4 paper

\IEEEoverridecommandlockouts	% This command is only needed if you want to use the \thanks command

\usepackage{times}

\usepackage{graphicx}
\graphicspath{{./Graphics/}}

\usepackage{subfigure}
\usepackage{tikz,pgf}
\usepackage{epsfig}

\usepackage{blindtext}

\usepackage{setspace}

\usepackage[utf8, latin1]{inputenc} % allow utf-8 input
\usepackage[T1]{fontenc}    		% use 8-bit T1 fonts
\usepackage{url}            		% simple URL typesetting
\usepackage{booktabs}       		% professional-quality tables
\usepackage{amsfonts}       		% blackboard math symbols
\usepackage{amsmath} 
\usepackage{amssymb}
\usepackage{nicefrac}       		% compact symbols for 1/2, etc.
\usepackage{microtype}      		% microtypography

\usepackage{tikz}
\usepackage{pgfplots}

\usepackage{tumabbrev}
\usepackage{xcolor}
\usepackage{siunitx}
\DeclareSIUnit{\px}{px}
\DeclareSIUnit{\Byte}{B}
\DeclareSIUnit{\fps}{fps}
\DeclareSIUnit{\bit}{bit}

\usepackage[breaklinks=true,letterpaper=true,colorlinks,bookmarks=false]{hyperref}

\title{\LARGE \bf 
Enhancing Traffic Scene Predictions with Generative Adversarial Networks
}

\author{Peter K\"onig$^{1}$, Sandra Aigner$^{1}$ and Marco K\"orner$^{1}$
\thanks{$^{1}$TUM Department of Civil, Geo and Environmental Engineering, Technical University of Munich, Germany
        {\tt\small \{peter.koenig, sandra.aigner, marco.koerner\}@tum.de}}%
}

\newcommand\copyrighttext{%
	\scriptsize \copyright~2019 IEEE. Personal use of this material is permitted. Permission from IEEE must be obtained for all other uses, in any current or future media, including reprinting/republishing this material for advertising or promotional purposes, creating new collective works, for resale or redistribution to servers or lists, or reuse of any copyrighted component of this work in other works.}%
\newcommand\copyrightnotice{%
	\begin{tikzpicture}[remember picture,overlay]
	\node[anchor=north,yshift=-20pt,xshift=0.3cm] at (current page.north) {{\parbox{\dimexpr\textwidth-\fboxsep-\fboxrule\relax}{\copyrighttext}}};
	\end{tikzpicture}%
}

\begin{document}

\maketitle
\copyrightnotice
\thispagestyle{empty}
\pagestyle{empty}

\begin{abstract} 
We present a new two-stage pipeline for predicting frames of traffic scenes where relevant objects can still reliably be detected. 
Using a recent video prediction network, we first generate a sequence of future frames based on past frames. 
A second network then enhances these frames in order to make them appear more realistic. 
This ensures the quality of the predicted frames to be sufficient to enable accurate detection of objects, which is especially important for autonomously driving cars. 
To verify this two-stage approach, we conducted experiments on the Cityscapes dataset. 
For enhancing, we trained two image-to-image translation methods based on generative adversarial networks, one for blind motion deblurring and one for image super-resolution. 
All resulting predictions were quantitatively evaluated using both traditional metrics and a state-of-the-art object detection network showing that the enhanced frames appear qualitatively improved.
While the traditional image comparison metrics, \ie MSE, PSNR, and SSIM, failed to confirm this visual impression, the object detection evaluation resembles it well. 
The best performing prediction-enhancement pipeline is able to increase the average precision values for detecting cars by about 9\% for each prediction step, compared to the non-enhanced predictions.
\end{abstract}

\section{Introduction}
Predicting possible future trajectories of objects in traffic scenes, such as cars and pedestrians, plays an essential role in anticipatory driving.
Only by having knowledge about the type of object and its possible movement patterns, we are able to make safe decisions as a human driver.
Having predictions as an additional input to a driver assistance system or an autonomous driving system would be beneficial to its internal decision-making process. 
Such a system could make faster and possibly more informed decisions regarding the control of the vehicle, which leads to an increase in safety.

Predicting the future frames of videos of street scenes is one way to anticipate the movement of objects. 
However, to support a system such as an autonomously driving car, the quality of the predicted frames must be high enough to enable the reliable detection of relevant objects.
Depending on the identified object, the decision process will vary greatly.
State-of-the-art object detection software produces good results on real videos of street scenes.
Thus, if a prediction looks as similar to the real data as possible, we can assume that detecting objects correctly will be easier.
\begin{figure}[t]
	\centering
	\vspace*{2mm}
	\includegraphics[width=.9\linewidth]{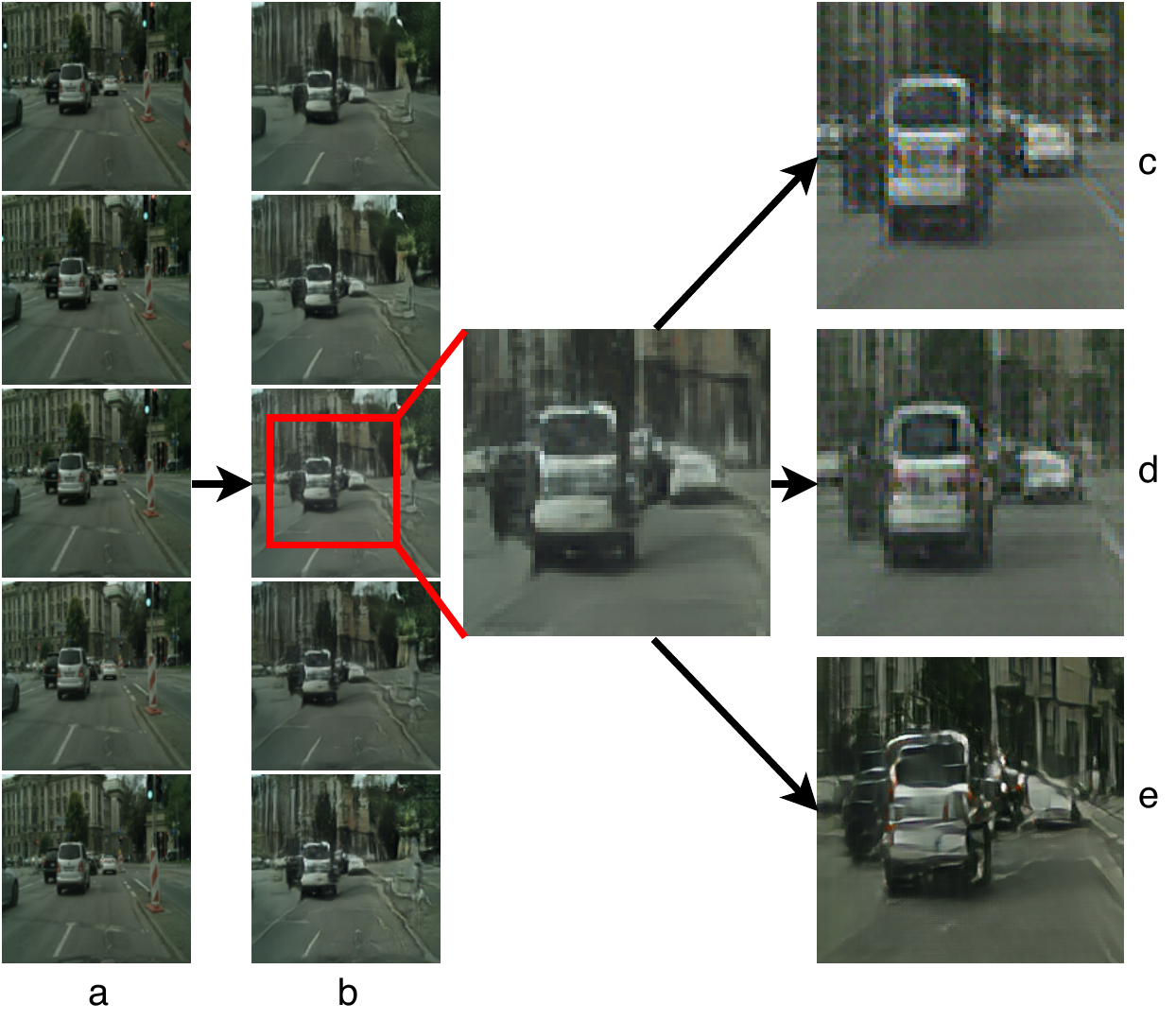}
	\vspace*{-1mm}
	\caption{Example predictions of the different prediction-enhancement pipelines. a: Input Sequence, b: FutureGAN \cite{Aigner2018}, c: DeblurGAN \cite{Kopyn2018} (transposed convolutions), d: DeblurGAN (NN-upsampling + convolution), e: SRGAN \cite{Ledig2016}.}
	\label{intro-example}
	\vspace*{-4mm}
\end{figure}

Due to the success of neural networks on a variety of computer vision tasks, we test the capabilities of neural network-based methods for generating enhanced video predictions that allow for the accurate detection of objects.
Particularly, we build on our previous video prediction network, \emph{FutureGAN} \cite{Aigner2018}, and predict five future frames of a street scene based on five input frames. 
The original results on the \emph{Cityscapes} dataset \cite{cordts2016} suggest that the network has learned reasonably good movement representations. 
However, for complex input data, such as natural street scenes, the predicted frames suffer from blurring effects and other unrealistic artifacts.
Therefore, we test several additional methods to enhance the predicted frames, thus making them appear more realistic, see figure \ref{intro-example} for example predictions. 
For enhancing, we utilize \emph{generative adversarial networks (GANs)} \cite{goodfellow2014}. 
In order to make the predictions more realistic and increase object detection results, we test two different GAN-based methods. 
The first one is an image super-resolution approach, the \emph{SRGAN} \cite{Ledig2016}, and the second one is a blind motion deblurring approach, the \emph{DeblurGAN} \cite{Kopyn2018}. 
In both cases, the frame enhancement is treated as an \emph{image-to-image translation} problem, where GANs have led to good results.

In this paper, we provide a reliable pipeline for predicting traffic scenes.
To prove the effectiveness of our prediction-enhancement pipeline, we evaluate all resulting predictions using the state-of-the-art object detection network \emph{YOLOv3} \cite{redmon2018}.
Our final model is able to produce predictions of both good visual quality and high detection accuracy.
The \emph{average precision (AP)} values for the object class "car" can be increased by about 9\% for each prediction step, compared to the non-enhanced predictions.

\section{Related Work}
Ranzato \etal \cite{Ranzato2014} first introduced a baseline for video prediction using deep neural networks. 
Since then, the deep learning-based prediction of future traffic sequences has become a widely researched topic in computer vision, especially in the autonomous driving community.

Due to the uncertainty in predicting the future, generating high-quality predictions of natural traffic scenes is a very complex task. 
This is why some approaches simplify this task and focus on predicting semantic segmentation masks, rather than generating the pixel values of the frames
\cite{Luc2017,Jin2017,Jin2017b,Nabavi2018,Bhattacharyya2019}. 
Many of these approaches, as well as approaches that directly generate the pixel values, use recurrent neural network structures \cite{DeBrabandere2016,Lotter2017,Elsayed2018,Wei2018,Kosiorek2018,Byeon2018,Nabavi2018,Xu2018}.
Lotter \etal \cite{Lotter2017}, for example, utilized long short-term memory (LSTM) \cite{Hochreiter1997} units to generate the pixel values of the frame one time step ahead.
Bhattacharjee \etal \cite{Bhattacharjee2017} generate predictions using a multi-stage GAN that takes input frames at different scales.
Using GANs \cite{Bhattacharjee2018,Aigner2018,Bhattacharjee2017}, or a combination of GANs and recurrent modules \cite{Liang2017}, are further common methods to predict video frames of traffic scenes.

Despite the recent advances in this field, the resulting frames often lack realism. 
To make predictions occur more realistic, others tackled the problem by learning separate representations for the static and dynamic components of a video. 
This is done either by incorporating motion conditions, such as optical flow information \cite{Gao2018,Reda2018,Hao2018,Jin2017,Liang2017}, or by learning sparse features that represent pixel dynamics \cite{Liu2018}.
Decomposing the video into static and non-static components allows the network to simply reproduce the values of the static part for the majority of pixels. 
Transformations are then only performed on the non-static pixels.
This leads to the problem of occluded and new objects not being properly modeled, especially in long-term predictions.  

Our approach builds on the idea of enhancing each prediction directly using a second network. 
Recently, related ideas without an application for traffic scenes were introduced \cite{Zhao2018,Xiong2018}. 
These approaches use two-stage networks to first generate subsequent frames from structure and content conditions, and then refine the frames using temporal signals or motion dynamics. 
We, on the other hand, use the learned motion representations of a GAN-based model to predict a set of future frames from a set of input frames. 
We then use a separate second model, an image-to-image translation GAN, to eliminate the artifacts and blurring effects caused by the transformations of the first network.

\section{Enhancement of Predicted Video Frames}
The methods used in this paper are based on GANs. 
In an adversarial setting, a generator network is trained to model the data distribution of the training data. 
During training, a second network, the discriminator, provides feedback to the generator about the similarity between the modeled and the observed data distribution. 
This results in a minimax game.
The discriminator $D$ tries to maximize its score of correctly classifying the samples it observes as real or fake.
The generator $G$ tries to fool the discriminator by minimizing the difference of the modeled and the data distribution, \ie by optimizing
\begin{equation}
\min_G \max_D \underset{x \sim \mathbb{P}_r}{\mathbb{E}} [\log (D(x))] + \underset{\tilde{x} \sim \mathbb{P}_g}{\mathbb{E}} [\log (1-D(\tilde{x}))] \; ,
\end{equation}
where $\mathbb{P}_r$ is the data distribution, $\mathbb{P}_g$ is the model distribution, implicitly defined by $\tilde{x} = G(z)$, and $z$ is the input sampled from a random distribution $P(z)$. 
During training, this approach gradually enforces the generator to produce samples that appear more and more similar to the training data.

However, there are problems with GAN-based approaches. 
First, they are hard to train and the highly unstable training process often leads to non-convergence.
Secondly, there is the mode collapse effect. 
This means, the generator learns to fool the discriminator by producing samples of a limited set of modes, thus produces samples that lack diversity. 
The generator fails to sufficiently model the variation in the real data distribution. 

For generating the traffic scene predictions we make use of our recent GAN-based approach, FutureGAN, that avoids these problems. 
We then evaluate how the predictions can be enhanced in order to improve the object detection results on the predicted frames.
The methods to enhance the predictions are all based on variants of the \emph{conditional GAN (cGAN)} \cite{mirza2014}, where enhancing is treated as an image-to-image translation problem. 
In the following, we describe the approaches used in this paper in more detail.

\subsection{FutureGAN}
To predict the future frames of the traffic sequences, we use FutureGAN. 
This network predicts multiple output frames from a set of input frames. 
It is trained using the \emph{progressively growing of GANs} technique, introduced by
Karras \etal \cite{karras2018}.
During training, layers are added progressively to both the generator and the discriminator network to increase the frame resolution gradually. 
Many architectures were particularly designed to overcome the GAN-related training issues, such as non-convergence and mode collapse \cite{Radford2016, Arjovsky2017b}. 
The progressive growing training strategy helps to further improve the GAN training. 
Additionally, the authors used feature normalization and a \emph{Wasserstein GAN with gradient penalty (WGAN-GP)} \cite{gulrajani2017} loss to increase the training stability of the network. 
For details on the network structure and architectural design, we refer the reader to the original paper \cite{Aigner2018}.

\subsection{DeblurGAN}
\label{sec:deblurgan}
As a first enhancement method, we chose DeblurGAN \cite{Kopyn2018}. 
DeblurGAN is a blind motion deblurring method based on cGANs. 
The DeblurGAN framework treats motion deblurring as an image-to-image translation problem. 
Rather than to estimate a motion kernel, the network is trained to directly translate the image from a blurry version to an unblurred one. 

The DeblurGAN loss function 
\begin{equation}
\label{eq:dgloss}
\mathcal{L} = \mathcal{L}_\text{GAN} + \lambda \mathcal{L}_\text{X}
\end{equation} 
consists of two components, 
a WGAN-GP loss term $\mathcal{L}_\text{GAN}$ and a content loss term $\mathcal{L}_\text{X}$, 
with $\lambda$ as a balancing factor. 
The content loss was introduced in addition to the adversarial loss term to increase the perceptual quality of the generated images.
In contrast to the standard $L_1$ \emph{(MAE)} or $L_2$ \emph{(MSE)} losses, which are based on the differences of the raw pixel values, this \emph{perceptual loss} \cite{Johnson2016} is based on the differences in feature space.
In particular,
\begin{equation}
\label{dgcontentloss}
\mathcal{L}_X = \frac{1}{W_{i,j} H_{i,j}} \displaystyle\sum_{x=1}^{W_{i,j}} \displaystyle\sum_{y=1}^{H_{i,j}} (\phi_{i,j}(I^S)_{x,y} - \phi_{i,j}(G_{\theta}(I^B))_{x,y})^2
\end{equation}
is the $L_2$ difference between the feature maps of the ground truth and the deblurred image of a specific layer in the VGG-19 \cite{simonyan2014} network, 
where $\phi_{i,j}$ is the feature map obtained before the $i$-th max-pooling layer and after the $j$-th convolutional layer of the VGG-19 network trained on ImageNet \cite{deng2009}, and $I^S$ and $I^B$ are the sharp ground truth and the blurry predicted frame. 
$W_{i,j}$ and $H_{i,j}$ are the width and height dimensions of the feature maps, respectively. 
In this case, we used the $VGG_{3,3}$ convolutional layer, since the general image content is typically captured in the lower layers of such a network \cite{zeiler2014}.

\begin{figure}
	\centering
	\includegraphics[width=0.85\linewidth]{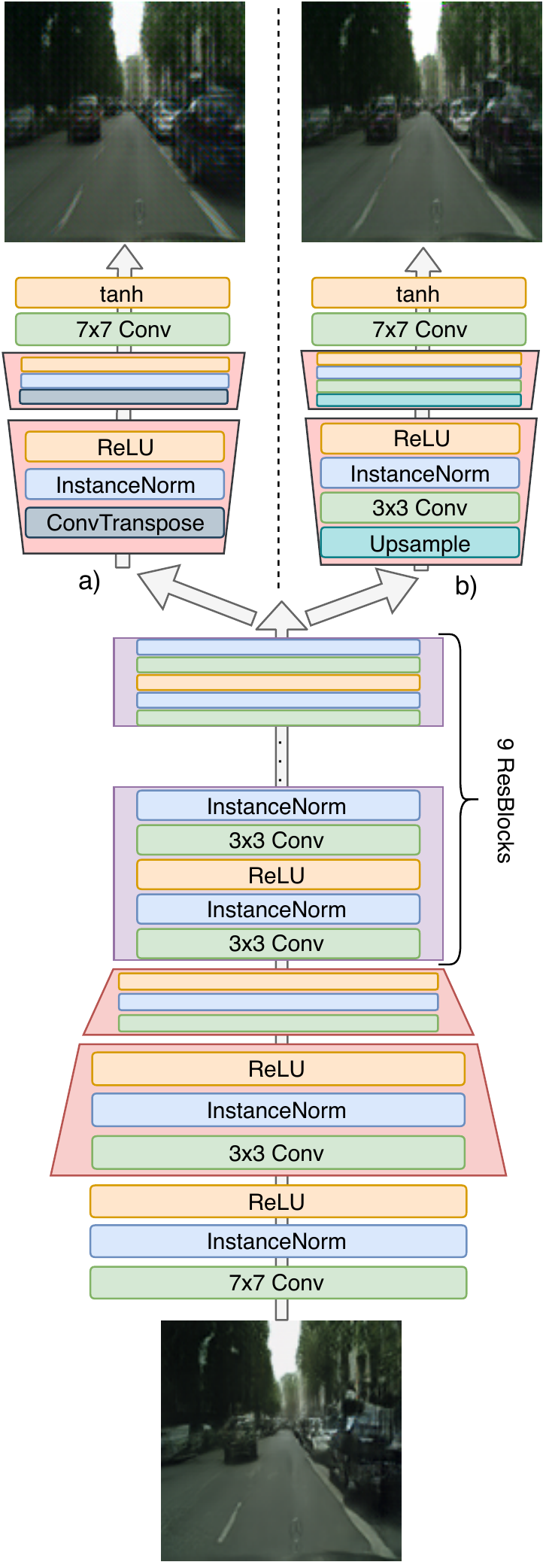}
	\caption{Generator structure of the two different DeblurGAN-based enhancement architectures. a) Regular structure as introduced by Kopyn \textit{et~al.} \cite{Kopyn2018}. b) Modified structure according to the suggestions of Odena \textit{et~al.} \cite{odena2016}.}
	\label{dg_structure}
\end{figure}

The original results of Kopyn \etal \cite{Kopyn2018} show that motion blur and artifacts similar to those of the FutureGAN street scene predictions can be removed effectively.  
After training DeblurGAN on our data, we observed that the network generates a checkerboard pattern on the deblurred test images (\cf Figure \ref{dg_structure}). 
Following the findings by Odena \etal \cite{odena2016}, we assume these patterns to be caused by the transposed convolutional layers in the upsampling part of the generator network. 
Transposed convolutions can produce this type of pattern because of the overlap that occurs when the kernel sizes are not divisible by the strides. 
To avoid such undesired patterns in the deblurred predictions, we designed a different version of DeblurGAN. 
We replaced each of the transposed convolutional layers in the original DeblurGAN architecture with a nearest-neighbor upsampling layer followed by a regular convolutional layer. 
The resulting generator structure can be seen in Figure \ref{dg_structure}. 
For completeness and comparability, we conducted separate experiments using both DeblurGAN versions.  

\subsection{SRGAN}
The second method we used to enhance the frames predicted by FutureGAN is a GAN-based approach for image super-resolution, the SRGAN \cite{Ledig2016}. 
Image super-resolution means that a low resolution (LR) image is upsampled to its high-resolution (HR) version. 
The SRGAN was designed to generate HR images of high perceptual quality, which are upsampled by a factor of 4 from the LR images.  
An increased resolution of the traffic scene predictions might also have positive effects on the object detection results because of the increased number of details in the high-resolution image.

The SRGAN loss is similar to the DeblurGAN loss (\cf Eq. \ref{eq:dgloss}). 
It also contains both an adversarial and a content loss term. 
The content loss is defined as in Eq. \ref{dgcontentloss} as the $L_2$ difference of the feature maps of a specific VGG-19 layer. 
In image super-resolution, the task is to recover high-frequency components, therefore
Ledig \etal \cite{Ledig2016} chose a deeper VGG-19 convolutional layer ($VGG_{5,4}$) to calculate the content loss. 
Similar as for DeblurGAN, the VGG-19 based content loss was chosen to learn perceptually meaningful representations for generating images of high visual quality. 
In the original experiments, SRGAN was able to recover high-level details in the images quite well and achieved high human rating based \emph{mean-opinion scores (MOS)}. 
For the detailed structure, we refer the reader to the original paper \cite{Ledig2016}.

\section{Experiments and Evaluation}
To evaluate the different enhancement methods for traffic scene prediction, the networks were trained on the Cityscapes dataset \cite{cordts2016}. 
This dataset consists of 30 frame long \SI{16}{\bit} color videos, which were recorded with a frame rate of \SI{17}{\fps} in 50 different German cities. 
The training split contains \num{2975} videos, the test split \num{1525}. 

For generating our initial predictions, we first trained FutureGAN according to the procedure described by Aigner and K\"orner \cite{Aigner2018}. 
The network was trained to predict five output frames from five input frames, thus the training and test sets contained \num{8924} and \num{4574} sequences, respectively.  
To avoid any overlap between the training and test split for all further experiments on the enhancement methods, we continued using only the Cityscapes test split as a database. 
We separated the new dataset into an 80:20 train-test split, leading to \num{3659} training sequences and \num{915} test sequences. 
The original input frames of size \SI{2048 x 1024}{\px} were downsampled bicubically to \SI{128 x 128}{\px} in all cases except for the ground truth frames for the SRGAN experiments, which were downsampled bicubically to \SI{512 x 512}{\px}. 
All networks are implemented in either PyTorch or Tensorflow for Python.

The training was performed on a single NVIDIA TITAN X Pascal GPU with \SI{12}{\giga\byte} of RAM separately for each network. 
We used the \emph{ADAM} optimizer \cite{kingma2015} for all networks. 
FutureGAN trained for 140 epochs with a gradually decaying learning rate of initially $l=0.001$ and $\beta_1 = 0.0$. 
Both DeblurGAN versions trained for 300 epochs with $\beta_1=0.5$ and an initial learning rate of $l=0.0001$ which gradually decayed to zero after 150 epochs. 
SRGAN was trained for 10 initialization epochs using the content loss and then for 300 full epochs with a learning rate of $0.0001$ and $\beta_1 = 0.9$.

After training, we evaluated the different prediction methods on our test split. 
For the plain predictions, we used the trained FutureGAN network to generate a set of five future frames from a set of five input frames. 
To test the different enhancement methods, we generated five predictions using FutureGAN and then enhanced each of the five frames using the different image-to-image translation networks.
In total, we evaluated four different prediction pipelines: plain FutureGAN (no enhancement), FutureGAN + DeblurGAN (transposed convolution), FutureGAN + DeblurGAN (upsample + convolution), and FutureGAN + SRGAN. 
In order to get an estimate of the inference time that it takes for predicting five future frames with each of the prediction pipelines, the following list provides the average values over the whole test set on an NVIDIA GeForce RTX 2070 GPU with \SI{8}{\giga\byte} of RAM:
\begin{itemize}
\item FutureGAN: \SI{0.011}{\second}
\item FutureGAN + DeblurGAN (trconv): \SI{0.019}{\second}
\item FutureGAN + DeblurGAN (ups+conv): \SI{0.020}{\second}
\item FutureGAN + SRGAN: \SI{0.707}{\second}
\end{itemize}
The SRGAN needs the most time to generate five predictions, most likely due to the increased frame size of the outputs.

\subsection{Qualitative Results}
\begin{figure*}
	\centering
	\includegraphics[width=\linewidth]{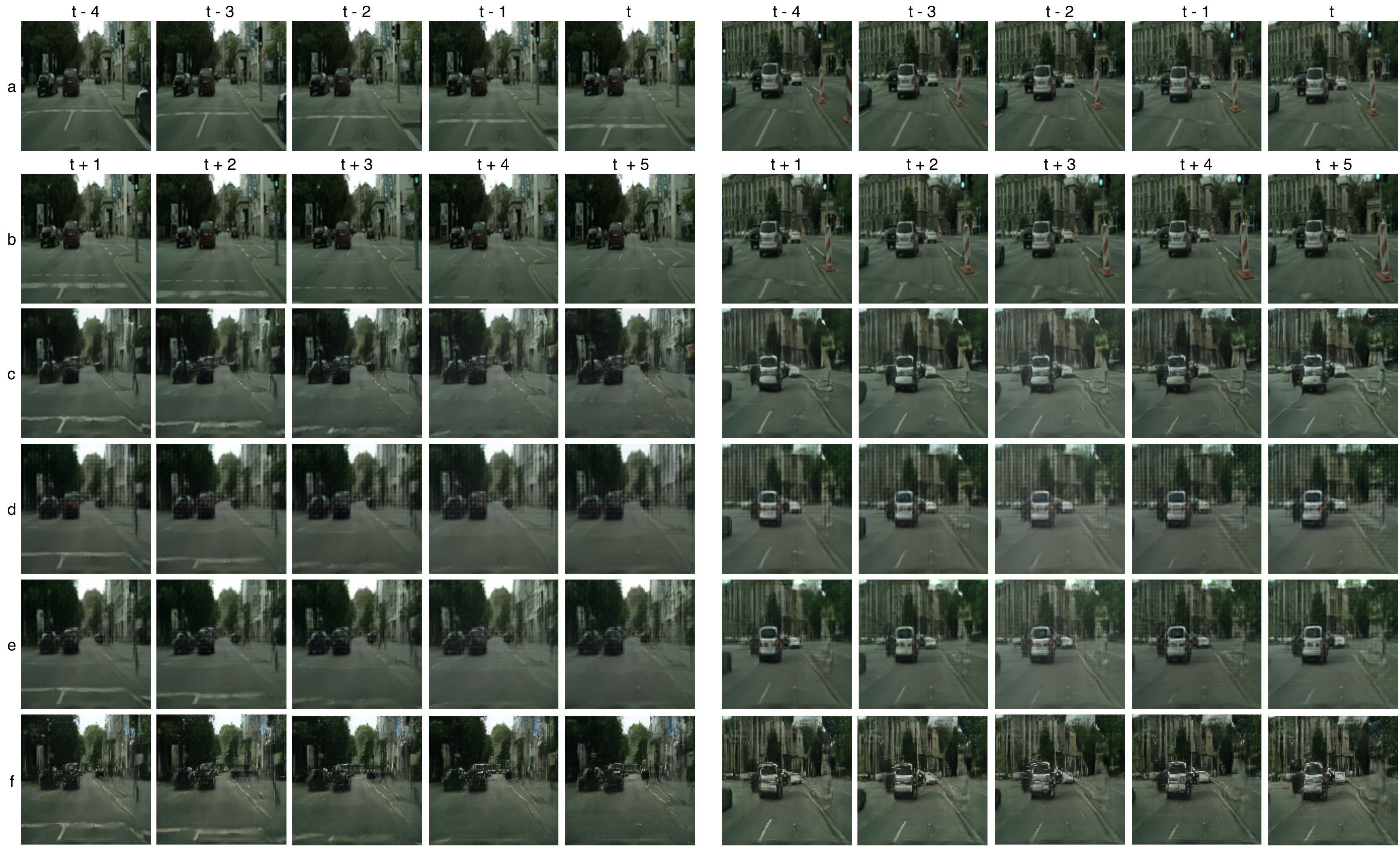}
	\caption{Prediction results for the Cityscapes test sequences. a: Input, b: Ground Truth, c: FutureGAN \cite{Aigner2018}, d: DeblurGAN \cite{Kopyn2018} (transposed convolutions), e: DeblurGAN (NN-upsampling + convolution), f: SRGAN \cite{Ledig2016}.}
	\label{cityscapes}
\end{figure*}

Figure \ref{cityscapes} shows a qualitative comparison of the prediction results for two different video sequences. 
For each of the two sequences, we display the input frames, the corresponding ground truth predictions, the prediction results of FutureGAN without any enhancement, and the results for the three different enhancement approaches. 
The differences between the enhancement methods are clearly visible. 
When using the original DeblurGAN architecture to enhance the predicted frames, the checkerboard pattern mentioned in section \ref{sec:deblurgan} can be seen in Figure \ref{cityscapes} d. 
Using the modified DeblurGAN with nearest-neighbor upsampling followed by regular convolutional layers reduces this pattern in the enhanced frames. 
In general, both DeblurGAN versions lead to an improved object appearance in all frames. 
Although there still remain unclear object boundaries after enhancing the frames, especially the cars and lane markings appear smoothed and straightened in comparison to the plain FutureGAN predictions. 
We further observed that the DeblurGAN architecture learned to generate object-specific features, such as the red colored taillights of cars (see figure \ref{cityscapes}). 
In contrast to that, the SRGAN-enhancement does not seem to produce a more realistic version of the predictions. 
The SRGAN learned to add high-frequency details to the image which do not match the original details.  This effect is probably caused by the content loss that is calculated with deeper VGG-19 feature maps. 
Even though SRGAN also generates object-specific features such as red taillights, the overall visual quality of the predicted frames seems best after the enhancement with the modified DeblurGAN (see Figure \ref{cityscapes} e).

\subsection{Quantitative Results: Traditional Metrics}
\begin{figure*}
	\centering
	\subfigure{
		\includegraphics[width=0.3\linewidth]{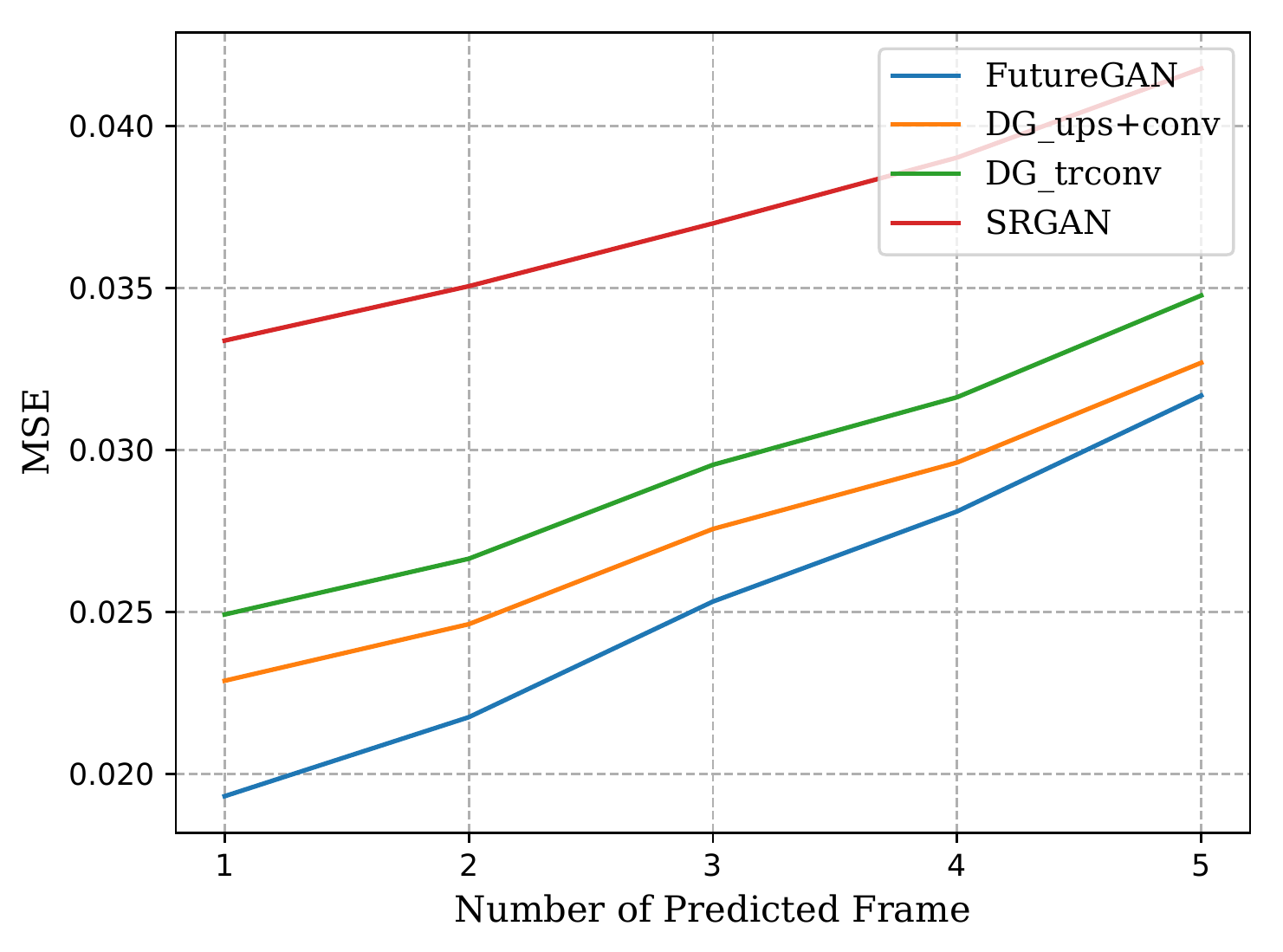}
		\label{mse}}
	\subfigure{
		\includegraphics[width=0.3\linewidth]{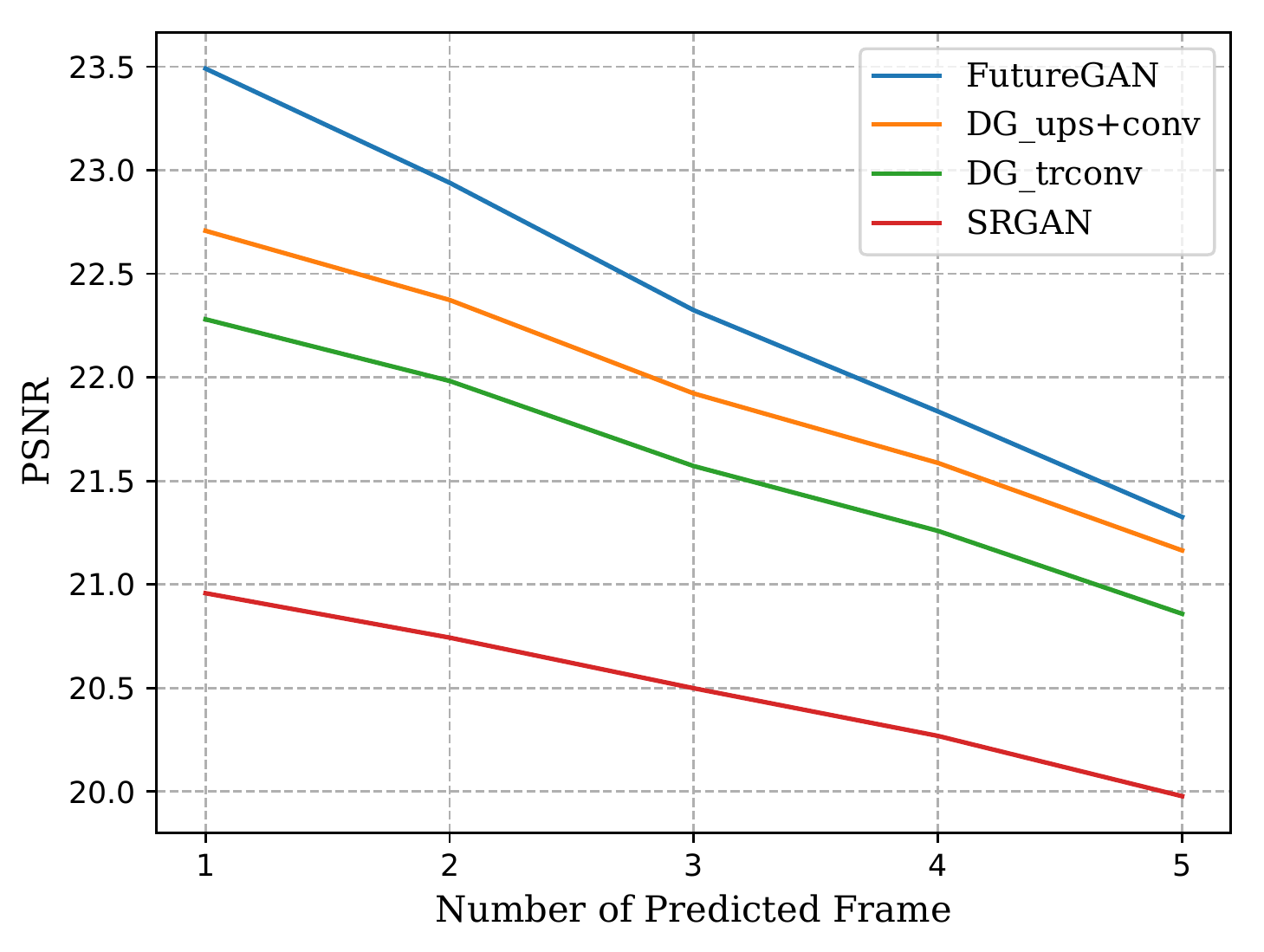}
		\label{psnr}}
	\subfigure{
		\includegraphics[width=0.3\linewidth]{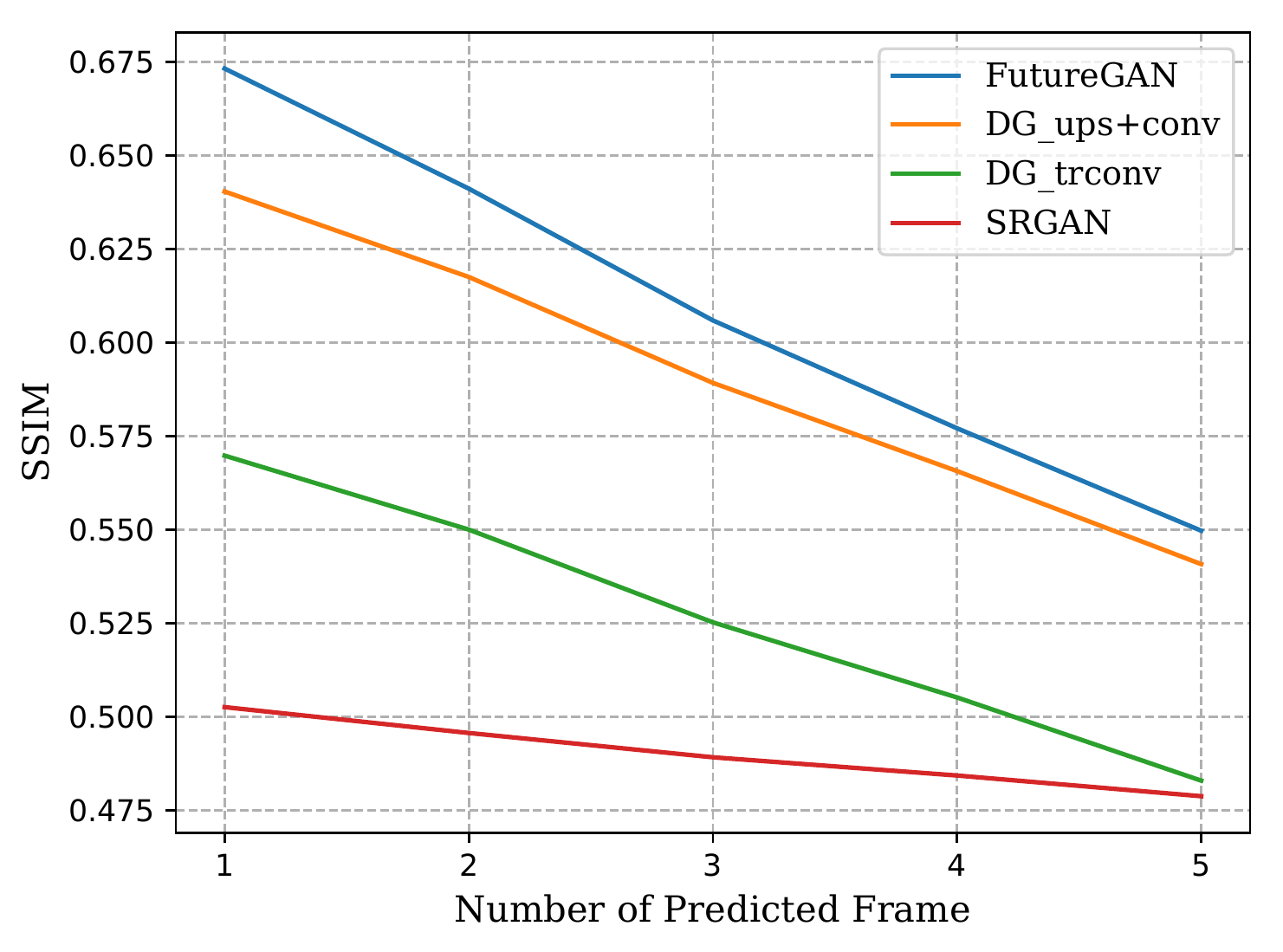}
		\label{ssim}}		
	\caption{Quantitative results per predicted frame for all enhancement methods (DG = DeblurGAN).}
	\label{quantitative}
\end{figure*}

A traditional way to quantitatively evaluate the enhanced predictions is to calculate image comparison metrics, such as the \emph{mean squared error (MSE)}, the \emph{peak signal-to-noise ratio (PSNR)}, and the \emph{structural similarity index (SSIM)}. 
For these evaluations, the resulting images are compared with the ground truth image of size \SI{128 x 128}{\px}, except for the case of SRGAN, where the comparison is on the increased size of \SI{512 x 512}{\px}.
The average values over all five frames are provided in Table \ref{table:metrics}. 
Additionally, we plotted the trends of the MSE, PSNR and SSIM values per predicted frame in figure \ref{quantitative}. 

\begin{table}
	\caption{Average results over 5 frames for all enhancement methods (best results in bold)}
	\label{table:metrics}
	\centering
	\begin{tabular}{lrrr}
		\cmidrule(r){1-4}
		& \multicolumn{1}{c}{MSE} & \multicolumn{1}{c}{PSNR} & \multicolumn{1}{c}{SSIM} \\
		\cmidrule(r){1-4}	
		FutureGAN \cite{Aigner2018}				& \textbf{0.0252} 	& \textbf{22.3829}	& \textbf{0.6094}\\
		DeblurGAN \cite{Kopyn2018} (trconv)		& 0.0295				& 21.5902    				& 0.5266\\
		DeblurGAN (ups+conv)					& 0.0275				& 21.9507    				& 0.5907\\
		SRGAN \cite{Ledig2016}					& 0.0372 				& 20.4893   				& 0.4900 \\
		\cmidrule(r){1-4}
	\end{tabular}
\end{table}

In general, the MSE, PSNR, and SSIM show worse results the higher the frame number. 
This was expected since a higher frame number represents a prediction further into the future. 
Looking in detail at the values, the non-enhanced FutureGAN predictions yield the best values (lower for MSE and higher for PSNR and SSIM). 
Enhancing the predictions using the modified DeblurGAN version (see Figure \ref{dg_structure} b) gives the second best results for all three metrics. 
When comparing the results of these traditional metrics, they seem contrary to the visual impression of the frames in figure \ref{cityscapes}. 
The traditional metrics can apparently not represent the human perception of improvement.
Figure \ref{cityscapes} shows this, when comparing the plain FutureGAN predictions (see Figure \ref{cityscapes} c) to the enhanced predictions of the modified DeblurGAN (see Figure \ref{cityscapes} e).

\subsection{Quantitative Results: Object Detection}
\begin{figure*}[t]
	\centering
	\includegraphics[width=.95\linewidth]{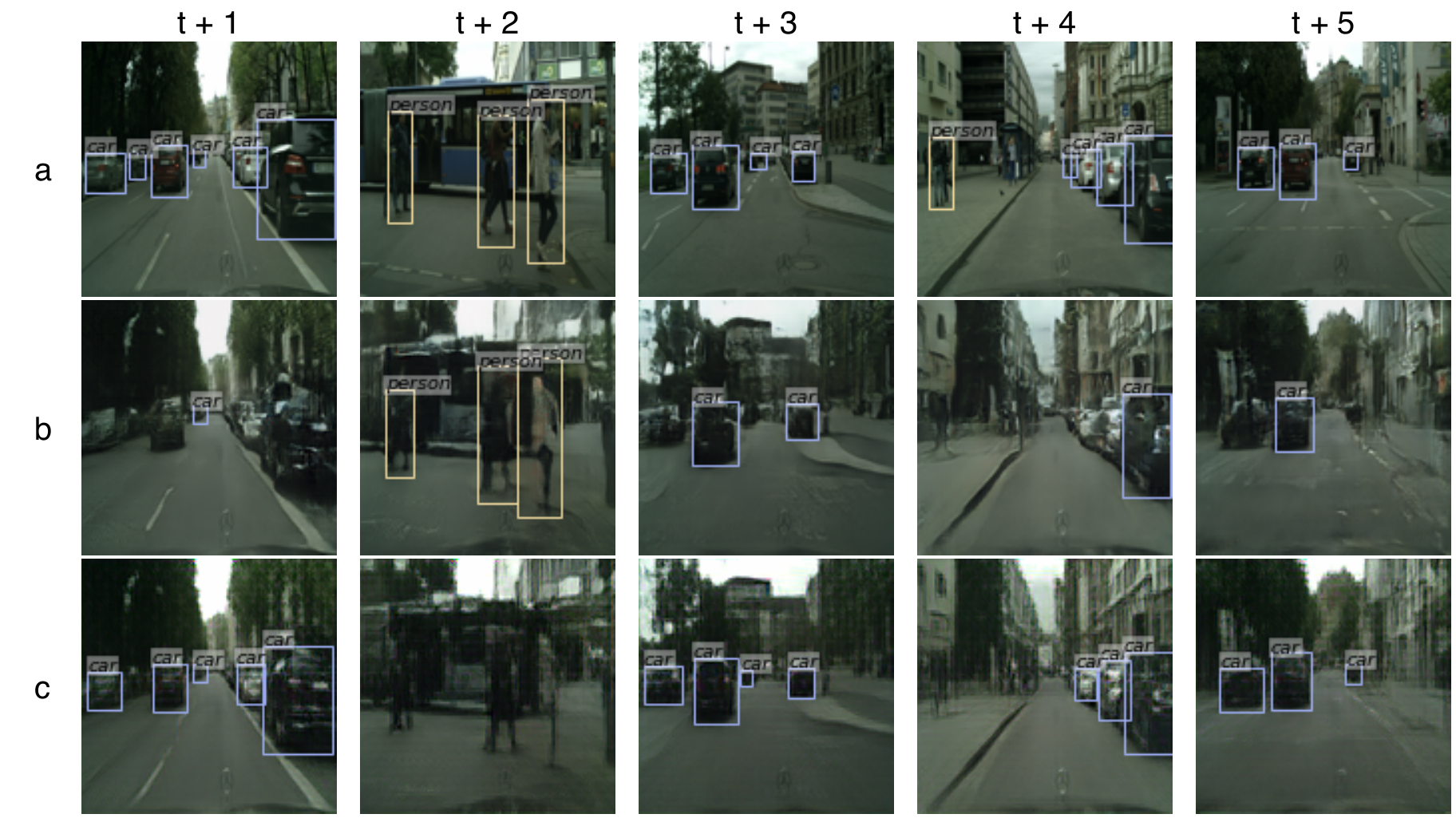}
	\caption{Object detection results of YOLOv3 \cite{redmon2018} from different video sequences at the five different time steps. a: Ground truth, b: plain FutureGAN \cite{Aigner2018}, c: DeblurGAN \cite{Kopyn2018} (upsample+convolution) enhanced prediction.}
	\label{Yolo}
\end{figure*}
To quantify the perceived visual improvement of the enhancement methods, especially also in the context of traffic scene prediction, we evaluated the images using a state-of-the-art object detection network, YOLOv3 \cite{redmon2018}. 
The network outputs bounding boxes and corresponding class labels. 
For evaluating the precision and recall of the object detection we take the detections on the ground truth frames as ground truth bounding boxes. 
This means the evaluation is relative with respect to the detection results of the algorithm on the ground truth images.

Figure \ref{Yolo} shows the qualitative results of the object detection network for four images. 
For brevity, we now only show the best performing enhancement method, the modified DeblurGAN (upsample + convolution), the plain FutureGAN predictions, and the ground truth frames. 
In these examples, one can see that, especially for the object class "car", the number of detections increases for the enhanced predictions in comparison with the non-enhanced predictions. 
However, the object detection network has problems detecting the class "person" in the enhanced images.
An example of this is also shown in figure \ref{Yolo}. 

\begin{figure}[h]
	\centering
	\includegraphics[width=0.9\linewidth]{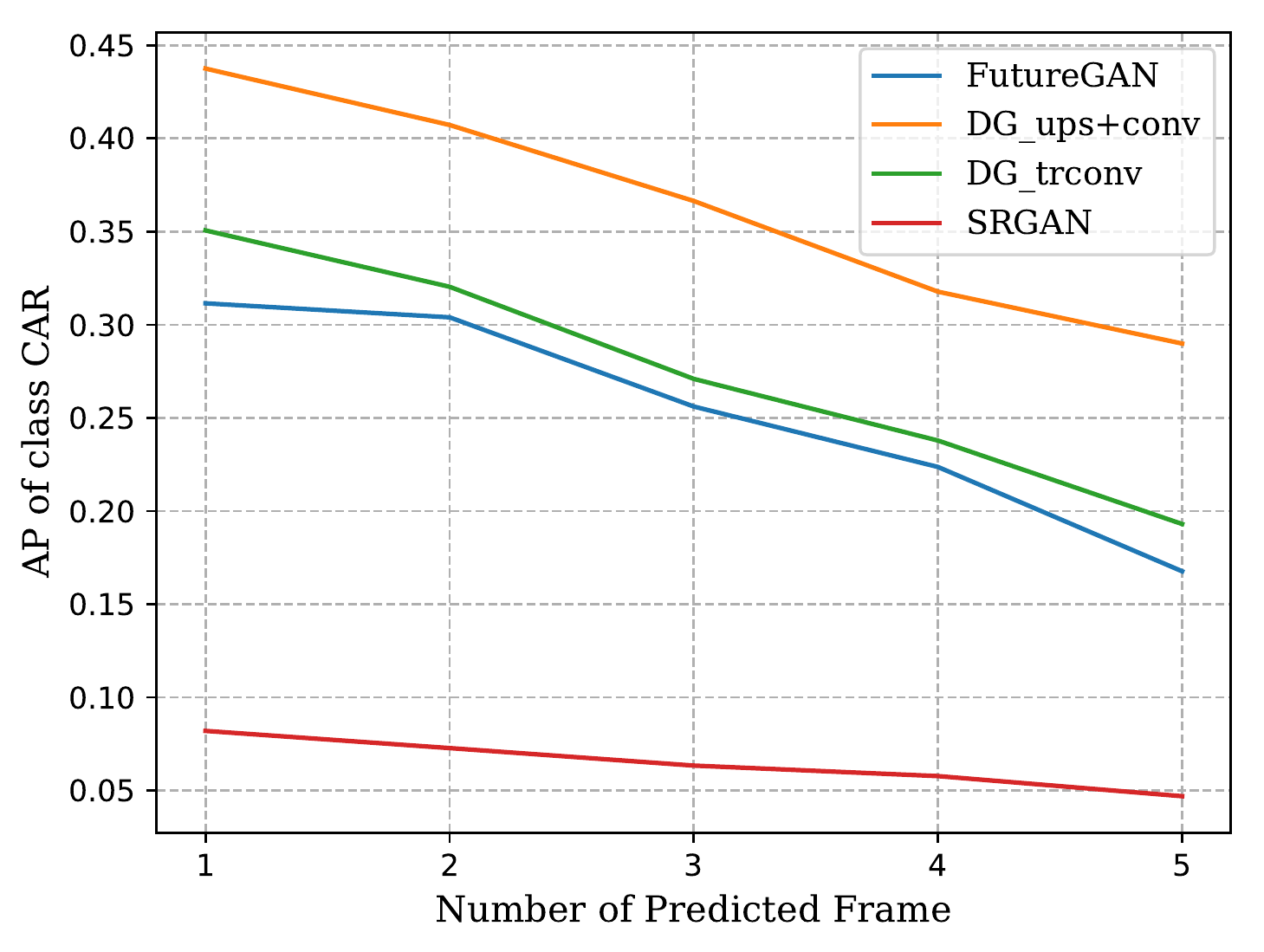}
	\caption{Average precision of the class "car" per predicted frame for all enhancement methods (DG = DeblurGAN).}
	\label{carAP}
\end{figure}
Since the class "car" is by far the most common class in our dataset, we specifically look at the average precision (AP) values of this class.
We calculate the AP as defined in \cite{everingham2010} with an IoU threshold of 50\% counting as correctly detected.  
Figure \ref{carAP} shows the development of the values per predicted frame for each of the prediction methods. 
For all methods, the general trend is a declining AP for an increased number of prediction steps, but the slow decrease suggests that cars are preserved well in the predicted frames.
Additionally, the qualitatively best performing enhancement method, DeblurGAN (upsample + convolution), shows the highest AP values for all five frames, which confirms the visual impression of the results.  
The SRGAN enhancement exhibits the lowest values throughout, which is also in accordance with the visual impression of the prediction results.

\section{Conclusions and Discussion}
In this paper, we evaluated the capabilities of GAN-based methods, SRGAN and DeblurGAN, to enhance video frame predictions of another generative model, FutureGAN. 
While, in general, motion representations and the difference between the movement of foreground and background objects are learned by FuturGAN, the predictions suffer from blurring effects on the moving objects, leading to irregular shapes and over-smoothed object details. 
In order to correct these effects, we used established image-to-image translation models to generate enhanced versions of the predicted frames. 
The networks were trained on the Cityscapes dataset to use them for traffic scene prediction.
We evaluated the different enhancement methods especially regarding their positive effects on object detection results, using a state-of-the-art object detector, the YOLOv3.

The visual quality of the enhanced predictions varies greatly between the enhancement methods. 
DeblurGAN shows a straightening effect, especially on car shapes and lane markings, leading to visually more realistic results.
Additionally, the network learns to include object-specific features such as car taillights, which initially were averaged out in the prediction results of FutureGAN. 
In contrast, SRGAN mainly learns to add high-frequency features to the enhanced image, which results in  unrealistic edges and patterns in the objects. 
These differences are most likely caused by the different content losses of DeblurGAN and SRGAN. 
While both networks use a very similar approach to calculate the content loss, DeblurGAN uses an earlier VGG-19 layer, SRGAN uses a deeper layer. 
With the low frame resolution in mind, a lower VGG-19 layer might be better for capturing the general content of the image. 

We evaluated the enhanced frame predictions using traditional image comparison metrics, but they failed to resemble the visual impression and showed no improvement for any of the enhancement methods. 
The evaluation of the object detection capabilities with YOLOv3, on the other hand, confirms the visual impression.
The enhancement method that produced the qualitatively best predictions, DeblurGAN (upsample + convolution), yielded the best detection performance. 
Even though pedestrians got over-smoothed by the enhancement network, possibly due to the low resolution of the images or the small number of training examples, the positive enhancement effect on cars is substantial. 
The average precision for detecting cars could be increased significantly in the enhanced predictions compared with the regular predictions.
This verifies the application of image-to-image translation models for enhancing predictions of traffic scenes. 

\bibliographystyle{ieee}
\bibliography{./Bib/Bib.bib}

\begin{thebibliography}{10}\itemsep=-1pt

\bibitem{Aigner2018}
S.~Aigner and M.~K\"{o}rner.
\newblock {F}uture{GAN}: {A}nticipating the {F}uture {F}rames of {V}ideo
  {S}equences using {S}patio-{T}emporal 3d {C}onvolutions in {P}rogressively
  {G}rowing {GAN}s.
\newblock {\em {T}he {I}nternational {A}rchives of the {P}hotogrammetry,
  {R}emote {S}ensing and {S}patial {I}nformation {S}ciences ({ISPRS})},
  XLII-2/W16:3--11, 2019.

\bibitem{Arjovsky2017b}
M.~Arjovsky, S.~Chitala, and L.~Bottou.
\newblock {W}asserstein {G}enerative {A}dversarial {N}etworks.
\newblock In {\em {ICML}}, volume~70, pages 214--223, 2017.

\bibitem{Bhattacharjee2017}
P.~Bhattacharjee and S.~Das.
\newblock {T}emporal {C}oherency based {C}riteria for {P}redicting {V}ideo
  {F}rames using {D}eep {M}ulti-stage {G}enerative {A}dversarial {N}etworks.
\newblock In {\em {N}eur{IPS}}. 2017.

\bibitem{Bhattacharjee2018}
P.~Bhattacharjee and S.~Das.
\newblock {C}ontext {G}raph based {V}ideo {F}rame {P}rediction using {L}ocally
  {G}uided {O}bjective.
\newblock In {\em {ECCV}: {W}orkshop on {A}nticipating {H}uman {B}ehavior},
  2018.

\bibitem{Bhattacharyya2019}
A.~Bhattacharyya, M.~Fritz, and B.~Schiele.
\newblock {B}ayesian {P}rediction of {F}uture {S}treet {S}cenes using
  {S}ynthetic {L}ikelihoods.
\newblock In {\em {ICLR}}, 2019.

\bibitem{Byeon2018}
W.~Byeon, Q.~Wang, R.~K. Srivastava, and P.~Koumoutsakos.
\newblock {F}ully {C}ontext-{A}ware {V}ideo {P}rediction.
\newblock In {\em {ECCV}}, 2018.

\bibitem{cordts2016}
M.~Cordts, M.~Omran, S.~Ramos, T.~Rehfeld, M.~Enzweiler, R.~Benenson,
  U.~Franke, S.~Roth, and B.~Schiele.
\newblock {T}he {C}ityscapes {D}ataset for {S}emantic {U}rban {S}cene
  {U}nderstanding.
\newblock In {\em {CVPR}}, pages 3213--3223, 2016.

\bibitem{DeBrabandere2016}
B.~De~Brabandere, X.~Jia, T.~Tuytelaars, and L.~Van~Gool.
\newblock {D}ynamic {F}ilter {N}etworks.
\newblock In {\em {N}eur{IPS}}. 2016.

\bibitem{deng2009}
J.~Deng, W.~Dong, R.~Socher, L.-J. Li, K.~Li, and L.~Fei-Fei.
\newblock {I}mage{N}et: {A} {L}arge-{S}cale {H}ierarchical {I}mage {D}atabase.
\newblock In {\em {CVPR}}, pages 248--255, 2009.

\bibitem{Elsayed2018}
N.~Elsayed, A.~S. Maida, and M.~Bayoumi.
\newblock {R}educed-{G}ate {C}onvolutional {LSTM} {U}sing {P}redictive {C}oding
  for {S}patiotemporal {P}rediction.
\newblock {\em {C}o{RR}}, abs/1810.07251, 2018.

\bibitem{everingham2010}
M.~Everingham, L.~Van~Gool, C.~K. Williams, J.~Winn, and A.~Zisserman.
\newblock {T}he {P}ascal {V}isual {O}bject {C}lasses ({VOC}) {C}hallenge.
\newblock {\em {I}nternational {J}ournal of {C}omputer {V}ision},
  88(2):303--338, 2010.

\bibitem{Gao2018}
H.~Gao, H.~Xu, Q.-Z. Cai, R.~Wang, F.~Yu, and T.~Darrell.
\newblock {D}isentangling {P}ropagation and {G}eneration for {V}ideo
  {P}rediction.
\newblock {\em {C}o{RR}}, abs/1812.00452, 2018.

\bibitem{goodfellow2014}
I.~Goodfellow, J.~Pouget-Abadie, M.~Mirza, B.~Xu, D.~Warde-Farley, S.~Ozair,
  A.~Courville, and Y.~Bengio.
\newblock {G}enerative {A}dversarial {N}etworks.
\newblock In {\em {NeurIPS}}, pages 2672--2680, 2014.

\bibitem{gulrajani2017}
I.~Gulrajani, F.~Ahmed, M.~Arjovsky, V.~Dumoulin, and A.~Courville.
\newblock {I}mproved {T}raining of {W}asserstein {GAN}s.
\newblock In {\em {NeurIPS}}, pages 5767--5777, 2017.

\bibitem{Hao2018}
Z.~Hao, X.~Huang, and S.~Belongie.
\newblock {C}ontrollable {V}ideo {G}eneration with {S}parse {T}rajectories.
\newblock In {\em {CVPR}}, 2018.

\bibitem{Hochreiter1997}
S.~Hochreiter and J.~Schmidhuber.
\newblock {L}ong {S}hort-{T}erm {M}emory.
\newblock {\em {N}eural {C}omputation}, 9(8):1735--1780, November 1997.

\bibitem{Jin2017}
X.~Jin, X.~Li, H.~Xiao, X.~Shen, Z.~Lin, J.~Yang, Y.~Chen, J.~Dong, L.~Liu,
  Z.~Jie, J.~Feng, and S.~Yan.
\newblock {V}ideo {S}cene {P}arsing {W}ith {P}redictive {F}eature {L}earning.
\newblock In {\em {ICCV}}, 2017.

\bibitem{Jin2017b}
X.~Jin, H.~Xiao, X.~Shen, J.~Yang, Z.~Lin, Y.~Chen, Z.~Jie, J.~Feng, and
  S.~Yan.
\newblock {P}redicting {S}cene {P}arsing and {M}otion {D}ynamics in the
  {F}uture.
\newblock In {\em {N}eur{IPS}}. 2017.

\bibitem{Johnson2016}
J.~Johnson, A.~Alahi, and L.~Fei-Fei.
\newblock {P}erceptual {L}osses for {R}eal-{T}ime {S}tyle {T}ransfer and
  {S}uper-{R}esolution.
\newblock In {\em {ECCV}}, 2016.

\bibitem{karras2018}
T.~Karras, T.~Aila, S.~Laine, and J.~Lehtinen.
\newblock {P}rogressive {G}rowing of {GAN}s for {I}mproved {Q}uality,
  {S}tability, and {V}ariation.
\newblock In {\em {ICLR}}, 2018.

\bibitem{kingma2015}
D.~P. Kingma and J.~Ba.
\newblock {A}dam: {A} {M}ethod for {S}tochastic {O}ptimization, 2015.

\bibitem{Kopyn2018}
O.~Kopyn, V.~Budzan, M.~Mykhailych, D.~Mishkin, and J.~Matas.
\newblock {D}eblur{GAN}: {B}lind {M}otion {D}eblurring {U}sing {C}onditional
  {A}dversarial {N}etworks.
\newblock In {\em {CVPR}}, 2018.

\bibitem{Kosiorek2018}
A.~R. Kosiorek, H.~Kim, I.~Posner, and Y.~W. Teh.
\newblock {S}equential {A}ttend, {I}nfer, {R}epeat: {G}enerative {M}odelling of
  {M}oving {O}bjects.
\newblock In {\em {N}eur{IPS}}. 2018.

\bibitem{Ledig2016}
C.~Ledig, L.~Theis, F.~Huszar, J.~Caballero, A.~Cunningham, A.~Acosta,
  A.~Aitken, A.~Tejani, J.~Totz, Z.~Wang, and W.~Shi.
\newblock {P}hoto-{R}ealistic {S}ingle {I}mage {S}uper-{R}esolution {U}sing a
  {G}enerative {A}dversarial {N}etwork.
\newblock In {\em {CVPR}}, 2016.

\bibitem{Liang2017}
X.~Liang, L.~Lee, W.~Dai, and E.~P. Xing.
\newblock {D}ual {M}otion {GAN} for {F}uture-{F}low {E}mbedded {V}ideo
  {P}rediction.
\newblock In {\em {ICCV}}, 2017.

\bibitem{Liu2018}
W.~Liu, A.~Sharma, O.~Camps, and M.~Sznaier.
\newblock {DYAN}: {A} {D}ynamical {A}toms-{B}ased {N}etwork {F}or {V}ideo
  {P}rediction.
\newblock In {\em {ECCV}}, 2018.

\bibitem{Lotter2017}
W.~Lotter, G.~Kreiman, and D.~Cox.
\newblock {D}eep {P}redictive {C}oding {N}etworks for {V}ideo {P}rediction and
  {U}nsupervised {L}earning.
\newblock In {\em {ICLR}}, 2017.

\bibitem{Luc2017}
P.~Luc, N.~Neverova, C.~Couprie, j.~Verbeek, and Y.~LeCun.
\newblock {P}redicting {D}eeper {I}nto the {F}uture of {S}emantic
  {S}egmentation.
\newblock In {\em {ICCV}}, 2017.

\bibitem{mirza2014}
M.~Mirza and S.~Osindero.
\newblock {C}onditional {G}enerative {A}dversarial {N}ets.
\newblock {\em {C}o{RR}}, abs/1411.1784, 2014.

\bibitem{Nabavi2018}
S.~S. Nabavi, M.~Rochan, and Y.~Wang.
\newblock {F}uture {S}emantic {S}egmentation with {C}onvolutional {LSTM}.
\newblock In {\em {BMVC}}, 2018.

\bibitem{odena2016}
A.~Odena, V.~Dumoulin, and C.~Olah.
\newblock {D}econvolution and {C}heckerboard {A}rtifacts.
\newblock {\em {D}istill}, 2016.

\bibitem{Radford2016}
A.~Radford, L.~Metz, and S.~Chintala.
\newblock {U}nsupervised {R}epresentation {L}earning with {D}eep
  {C}onvolutional {G}enerative {A}dversarial {N}etworks.
\newblock In {\em {ICLR}}, 2016.

\bibitem{Ranzato2014}
M.~Ranzato, A.~Szlam, J.~Bruna, M.~Mathieu, R.~Collobert, and S.~Chopra.
\newblock {V}ideo ({L}anguage) {M}odeling: {A} {B}aseline for generative
  {M}odels of natural {V}ideos.
\newblock {\em {C}o{RR}}, abs/1412.6604, 2014.

\bibitem{Reda2018}
F.~A. Reda, G.~Liu, K.~J. Shih, R.~Kirby, J.~Barker, D.~Tarjan, A.~Tao, and
  B.~Catanzaro.
\newblock {SDC}-{N}et: {V}ideo prediction using spatially-displaced
  convolution.
\newblock In {\em {ECCV}}, 2018.

\bibitem{redmon2018}
J.~Redmon and A.~Farhadi.
\newblock {YOLO}v3: {A}n {I}ncremental {I}mprovement.
\newblock {\em {C}o{RR}}, abs/1804.02767, 2018.

\bibitem{simonyan2014}
K.~Simonyan and A.~Zisserman.
\newblock {V}ery {D}eep {C}onvolutional {N}etworks for {L}arge-{S}cale {I}mage
  {R}ecognition.
\newblock In {\em {ICLR}}, 2015.

\bibitem{Wei2018}
H.~Wei, X.~Yin, and P.~Lin.
\newblock {N}ovel {V}ideo {P}rediction for {L}arge-scale {S}cene using
  {O}ptical {F}low.
\newblock {\em {C}o{RR}}, abs/1805.12243, 2018.

\bibitem{Xiong2018}
W.~Xiong, W.~Luo, L.~Ma, W.~Liu, and J.~Luo.
\newblock {L}earning to {G}enerate {T}ime-{L}apse {V}ideos {U}sing
  {M}ulti-{S}tage {D}ynamic {G}enerative {A}dversarial {N}etworks.
\newblock In {\em {CVPR}}, 2018.

\bibitem{Xu2018}
J.~Xu, B.~Ni, Z.~Li, S.~Cheng, and X.~Yang.
\newblock {S}tructure {P}reserving {V}ideo {P}rediction.
\newblock In {\em {CVPR}}, 2018.

\bibitem{zeiler2014}
M.~D. Zeiler and R.~Fergus.
\newblock {V}isualizing and {U}nderstanding {C}onvolutional {N}etworks.
\newblock In {\em {ECCV}}, pages 818--833. Springer, 2014.

\bibitem{Zhao2018}
L.~Zhao, X.~Peng, Y.~Tian, M.~Kapadia, and D.~Metaxas.
\newblock {L}earning to {F}orecast and {R}efine {R}esidual {M}otion for
  {I}mage-to-{V}ideo {G}eneration.
\newblock In {\em {ECCV}}, 2018.

\end{thebibliography}

\end{document}